\documentclass[letterpaper, 10 pt, conference]{ieeeconf}  

\IEEEoverridecommandlockouts                              
\usepackage{dirtytalk}

\usepackage{graphicx}
\usepackage{mathtools}
\usepackage{amssymb}

\usepackage[english]{babel}
\usepackage{amsthm}
\theoremstyle{definition}
\newtheorem{definition}{Definition}

\usepackage{caption}
\usepackage{subcaption}
\usepackage{floatrow}

\let\vec\mathbf

\usepackage{booktabs}

\usepackage{multirow}

\title{\LARGE \bf
LEAFAGE: Example-based and Feature importance-based Explanations for Black-box ML models
}

\author{Ajaya Adhikari$^{1}$, David M. J. Tax$^{2}$, Riccardo Satta$^{1}$, Matthias Faeth$^{2}$
\thanks{$^{1}$TNO, The Hague, The Netherlands}%
\thanks{$^{2}$Delft University of Technology, Delft, The Netherlands}%
}

\begin{document}

\maketitle
\thispagestyle{empty}
\pagestyle{empty}

\begin{abstract}
Explainable Artificial Intelligence (XAI) is an emergent research field which tries to cope with the lack of transparency of AI systems, by providing human understandable explanations for the underlying Machine Learning models.
This work presents a new explanation extraction method called LEAFAGE.
Explanations are provided both in terms of feature importance and of similar classification examples. The latter is a well known strategy for problem solving and justification in social science. 
LEAFAGE leverages on the fact that the reasoning behind a single decision/prediction for a single data point is generally simpler to understand than the complete model; it produces explanations by generating simpler yet locally accurate approximations of the original model.
LEAFAGE performs overall better than the current state of the art in terms of fidelity of the model approximation, in particular when Machine Learning models with non-linear decision boundaries are analysed.
LEAFAGE was also tested in terms of usefulness for the user, an aspect still largely overlooked in the scientific literature. Results show interesting and partly counter-intuitive findings, such as the fact that providing no explanation is sometimes better than providing certain kinds of explanation.
\end{abstract}


\section{Introduction}
In the context of Artificial Intelligence, Machine Learning (ML) is a rapidly growing field.
There has been a surge of high-performance models for classification and prediction.
Still, the application of these models in high-risk domains is more stagnant due to lack of transparency and trust: there is a disconnect between the black-box character of these models and the needs of the users.
Explainable Artificial Intelligence (XAI) has recently emerged to provide solutions to this issue by attempting to create understandable explanations for the reasoning of a black-box model. 

Example-Based Reasoning (EBR), i.e., motivating a decision by providing examples of similar situations, is widely recognized as an effective way to provide explanations \cite{aamodt1994case}, as it bears a close resemblance to the way humans think. 
As a result, it is commonly used e.g. in the health-care sector for decision-support systems \cite{begum2011case, bichindaritz2006case} and in law for justifying arguments, positions and decisions \cite{kolodner1992introduction}.
However, the usage of EBR to explain black-box ML models (i.e., models whose inner mechanisms are either unknown by the user, or too complex to be practically comprehensible by a human), has been largely overlooked so far in the scientific literature. 
This is partly because of the difficulty of finding examples according to the inner reasoning of such a model. 
Notably, most of the scientific literature focuses instead on evaluating the relative importance of features (\textit{feature importance-based} explanations, see e.g. LIME \cite{ribeiro2016should}).

In this paper, we propose a new method for providing both feature importance-based and EBR explanations of the local reasoning of black-box models. 
Here, \textit{local} refers to the ability of tailoring the explanation to a single prediction taken by the ML model, as opposed to providing a global explanation of the whole model logic. 
We named the method \textit{LEAFAGE} - Local Example and Feature importance-based model AGnostic Explanations.
LEAFAGE approximates the local reasoning of the black-box model by a (transparent) linear model.
As a byproduct, LEAFAGE is also able to provide the importance of each feature for a prediction.

We evaluate LEAFAGE both in terms of fidelity, and of usefulness to the user.
\textit{Fidelity} refers to whether the extracted explanation reflects the true reasoning of the underlying black-box ML model.
The \textit{usefulness} to the user is evaluated by conducting a user-study in terms of perceived aid in decision-making and objective transparency.

The remainder of this paper is structured as follows.
In Chapter \ref{section:background}, we provide background information and related work on XAI, and explore approaches to provide explanations that leverage on social research.
Chapter \ref{section:leafage} describes LEAFAGE, which is then evaluated in terms of fidelity and usefulness to the user in Chapters \ref{chapter:quantitative_evaluation} and \ref{chapter:empirical_evaluation}.
Finally, Chapter \ref{chapter:conclusion} draws conclusions and suggests future research directions.

\section{Background}\label{section:background}
This Section surveys current literature on XAI for ML models, and on the user's perspective on an explanation.

An explanation about a ML model can be of \textit{global} or \textit{local} scope.
A \textit{global} explanation clarifies the inner workings of the whole ML model, i.e., how the relationship between input and output spaces is modeled \cite{gill2018introduction}.
\textit{Local} explanations look instead at the reasoning behind a decision/prediction over a single input data point (\textit{test} sample), thus targeting a sub-region of the input space.
As the complexity of the ML model grows, it becomes harder to generate an understandable global explanation. 
However, it is likely that the logic of the ML model in the neighbourhood of a single test sample will be much simpler, thus allowing to generate understandable \textit{local} explanations.

Three main strategies for extracting human-understandable explanations from ML models can be found in the literature: \textit{transparent-by-design}, \textit{model-oriented} and \textit{model-agnostic}.
In the first strategy, the ML model is designed from the start to be globally transparent and possibly simple enough to be understandable by humans (e.g. a small decision tree).
The latter two strategies deal instead with an existing model that has not been made transparent by design. 
In the model-oriented strategy, certain parts of the model are used to extract an explanation (e.g., see \cite{xu2015show}).
In case when the ML model is too complex, or when internal workings of the model are not accessible, a model-agnostic strategy is used.
This strategy views the ML model as a \textit{black-box}, and queries it using a set of instances from the input space in order to gain insights in the behaviour of the model.

\begin{figure}
\captionsetup{justification=centering}
\begin{center}
\includegraphics[width=0.5\textwidth]{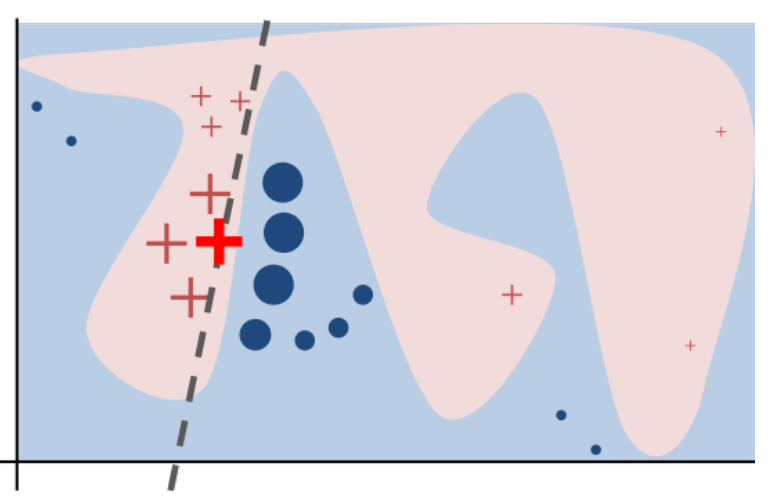}%
\caption{The dashed line approximates the blue/red  decision boundary in the neighbourhood of the bold `+' point \cite{ribeiro2016should}.}
\label{fig:lime_sampling}
\end{center}
\end{figure}

The proposed method, LEAFAGE, falls into the latter category.
One of the most recent methods on the same category is LIME (Local Interpretable Model-agnostic Explanations) \cite{ribeiro2016should}, from which LEAFAGE borrows its main ideas. 
LIME provides a \textit{local explanation} by linearly approximating the decision boundary of the ML model in the neighbourhood of the test sample.
Figure \ref{fig:lime_sampling} shows an example of how LIME works in a binary classification problem. 
The two classes are Red and Blue, respectively represented by '+' and full circles.
The decision boundary of the ML model is between yellow/blue areas. 
The point marked with a bold '+' is the test sample $\vec{z}$. In order to generate an explanation as for why the model ML classified $\vec{z}$ as Red, other Red and Blue synthetic data points are sampled from the input space.

A linear model is learned on the synthetic data points; higher importance is given in correctly classifying the synthetic instances that are close to $\vec{z}$.
In Figure \ref{fig:lime_sampling} their size represents their proximity to $\vec{z}$.
This proximity from an synthetic instance $\vec{x}$ to $\vec{z}$ is defined by an exponential kernel $\pi(\vec{z}, \vec{x}) = e^{-D(\vec{x},\vec{z})^2/\sigma^2}$. In \cite{ribeiro2016should}, $\sigma$ is fixed to $0.75*\sqrt{dimension}$. 

The parameter $\sigma$ of the proximity kernel plays an important role: a fixed $\sigma$ can lead to neighbourhoods that do not include a decision boundary, or that include a too big part of the decision boundary which cannot be approximated linearly. 
Laugel et al. \cite{laugel2018defining} spotted this problem and suggested to sample instances close to the nearest decision boundary to $\vec{z}$ within a fixed hyper-sphere.
However, the fixed hyper-sphere can also lead to too small or too big neighbourhoods.

XAI research has been criticized for overlooking the viewpoint of the end-user, i.e., if he/she is satisfied with provided explanations \cite{miller2017explainable}.
To address this, the present work focuses on providing explanations according to Example-Based Reasoning (EBR), a paradigm where explanations are related to previous experience \cite{kolodner1992introduction, richter2016case}.
This type of reasoning lies very close to how humans think \cite{bichindaritz2006case, aamodt1994case}.

EBR applications can be divided into two types: \textit{problem-solving} and \textit{decision-justification} \cite{kolodner1992introduction}.
In \textit{problem-solving}, previous similar situations are used as aid to decide how to proceed with the current situation.
In \textit{decision-justification}, previous similar situations are leveraged to support or dismiss certain arguments and decisions.
Worth noting, Common Law, which is used in most English-speaking countries, is based on the same principle (judicial decision are made on similar cases from the past \cite{richter2016case}).

The ultimate goal of an ML explanation system is to provide valuable insights on an automated prediction/decision to the user. Such aspect can be evaluated by conducting user-studies.
In recommendation systems, extensive research has been conducted in designing user-studies which evaluate explanations that clarify why a certain item is recommended, from the user's point of view 
\cite{tintarev2011designing, gedikli2014should, cramer2008effects}.
Tintarev \cite{tintarev2011designing} defines seven goals for an explanation system, namely \textit{transparency}, \textit{scrutability}, \textit{trust}, \textit{effectiveness}, \textit{efficiency}, \textit{persuasiveness} and \textit{satisfaction}.
All of them can be evaluated subjectively by asking questions to the user \cite{tintarev2011designing}.
However, while \textit{trust}, \textit{satisfaction} and \textit{effectiveness} are subjective by nature, \textit{transparency}, \textit{efficiency} and \textit{persuasiveness} can also be measured objectively. E.g., one can test whether users have understood the reasoning behind the recommendations \cite{cramer2008effects}, measure the interaction time \cite{gedikli2014should} or check whether the user agrees to buy a recommended item.


\section{LEAFAGE}\label{section:leafage}
This Section describes our proposed method, LEAFAGE (Local Example and Feature importance-based model AGnostic Explanation). 
It provides explanations in the form of examples drawn from the training set, that are similar to the test sample \textit{according to the ML model logic}, and shows the importance of each feature for the prediction.

Let $f: \mathcal{X} \rightarrow \mathcal{Y}$ be a black-box ML model that solves a binary classification problem with $\mathcal{X}=\mathbb{R}^d$ and $\mathcal{Y}=\{c_1, c_2\}$, and $\vec{z} \in \mathcal{X}$ be the test sample, an instance of the input space with $f(\vec{z})=c_z$, $c_z \in \mathcal{Y}$.

Furthermore, let $X=[\vec{x}_1,..,\vec{x}_n]$ with the corresponding true labels $y_{true}=[y_1,..,y_n]$ be the training set used to train $f$, and $y_{predicted}=\{f(\vec{x}_i) | \vec{x}_i \in X\}$ be the predicted labels of the training set.
Next, let $\{\vec{x} \in \mathcal{X}\ | f(\vec{x}) = c_z\}$ and $\{\vec{x} \in \mathcal{X}\ | f(\vec{x}) \neq c_z\}$ be defined as the \textit{ally} and the \textit{enemy} instances of $\vec{z}$ \cite{laugel2018comparison}, respectively.

LEAFAGE uses $X$, $y_{predicted}$, $\vec{z}$ and $c_z$ to explain why $\vec{z}$ was predicted as $c_z$.
It works as follows:
\begin{itemize}[*]
    \item A subset of the training set in the neighbourhood of $\vec{z}$ is used to build a local linear model. The coefficients of this model provide a measure of importance of each feature locally.
    \item These coefficients are used to define a local dissimilarity measure between any instance $\vec{x_i} \in \mathcal{X}$ and $\vec{z}$. In turn, this measure is used to retrieve examples similar to $\vec{z}$ from the training set.
    \item The importance of each feature and the most similar examples are given as  explanation of the classification.
\end{itemize}

Section \ref{section:defining_dissimilarity} and \ref{section:computation_local_linear_model} illustrate respectively the adopted dissimilarity measure, and the strategy to build the local linear model.
Next, Section \ref{section:explanation_extraction} explains how LEAFAGE explanations can be presented to the user.

\subsection{Defining a local dissimilarity measure}\label{section:defining_dissimilarity}
\begin{figure}[b]
\captionsetup[subfigure]{justification=centering}
\captionsetup{justification=centering}
    \centering
    \begin{subfigure}[b]{0.30\textwidth}
        \centering
        \includegraphics[width=\textwidth]{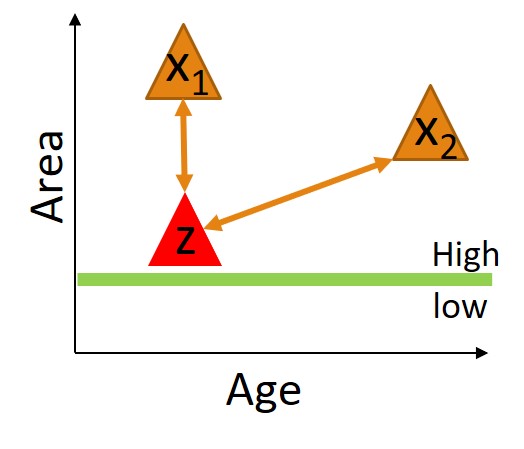}
        \caption{Euclidean distance: $\vec{z}$ is more similar to $\vec{x}_1$ than $\vec{x}_2$}
        \label{fig:simple_boundary_euclidean_distance}
    \end{subfigure}
    \hfill
    \begin{subfigure}[b]{0.30\textwidth}
        \centering
        \includegraphics[width=\textwidth]{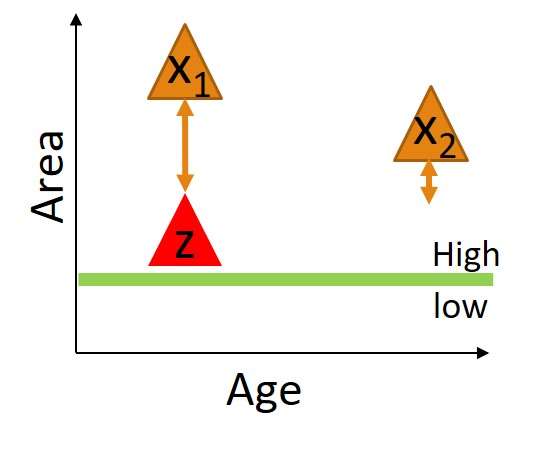}
        \caption{Black-box classifier: $\vec{z}$ is more similar to $\vec{x}_2$ than $\vec{x}_2$}
        \label{fig:simple_boundary_black_box_distance}
    \end{subfigure}
    \hfill
    \begin{subfigure}[b]{0.30\textwidth}
        \centering
        \includegraphics[width=\textwidth]{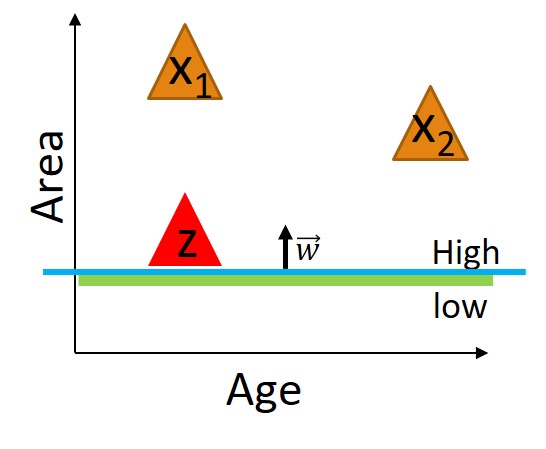}
        \caption{Approximate the decision boundary with a linear model.}
        \label{fig:simple_boundary_linear_model}
    \end{subfigure}
    \caption{Illustration of different types of distances.}
    \label{fig:simple_boundary_distances}
\end{figure}

Consider a binary classification problem where an ML model predicts whether a house has a \textit{high} or \textit{low} value according to two features, \textit{area} and \textit{age}, as shown in Figure \ref{fig:simple_boundary_euclidean_distance} (in green, the decision boundary of a simple linear classifier).
A test house $\vec{z}$ is predicted as value \textit{high}. To find similar houses in the training set, one could use the Euclidean distance (Figure \ref{fig:simple_boundary_euclidean_distance}).
However, this choice does not reflect the reasoning of the classifier, which only looks at the feature \textit{area}; in fact, according to the classifier, $\vec{z}$ is more similar to $\vec{x}_2$ than $\vec{x}_1$ (Figure \ref{fig:simple_boundary_black_box_distance})


A way to compute a dissimilarity measure that takes into account the reasoning of the classifier is to use feature weights derived from a local linear approximation $\hat f_z(\vec{x})= \vec{w}_z \vec{x} + c$ with $\vec{w}_z=(w_{z1}, ..., w_{zd})^T$ of the decision boundary (the blue line in Figure \ref{fig:simple_boundary_linear_model}).
Then, $\vec{w}_z$ will denote the most discriminative direction for the classification of $\vec{z}$.

\begin{figure}
\captionsetup{justification=centering}
\begin{center}
\includegraphics[width=1\textwidth]{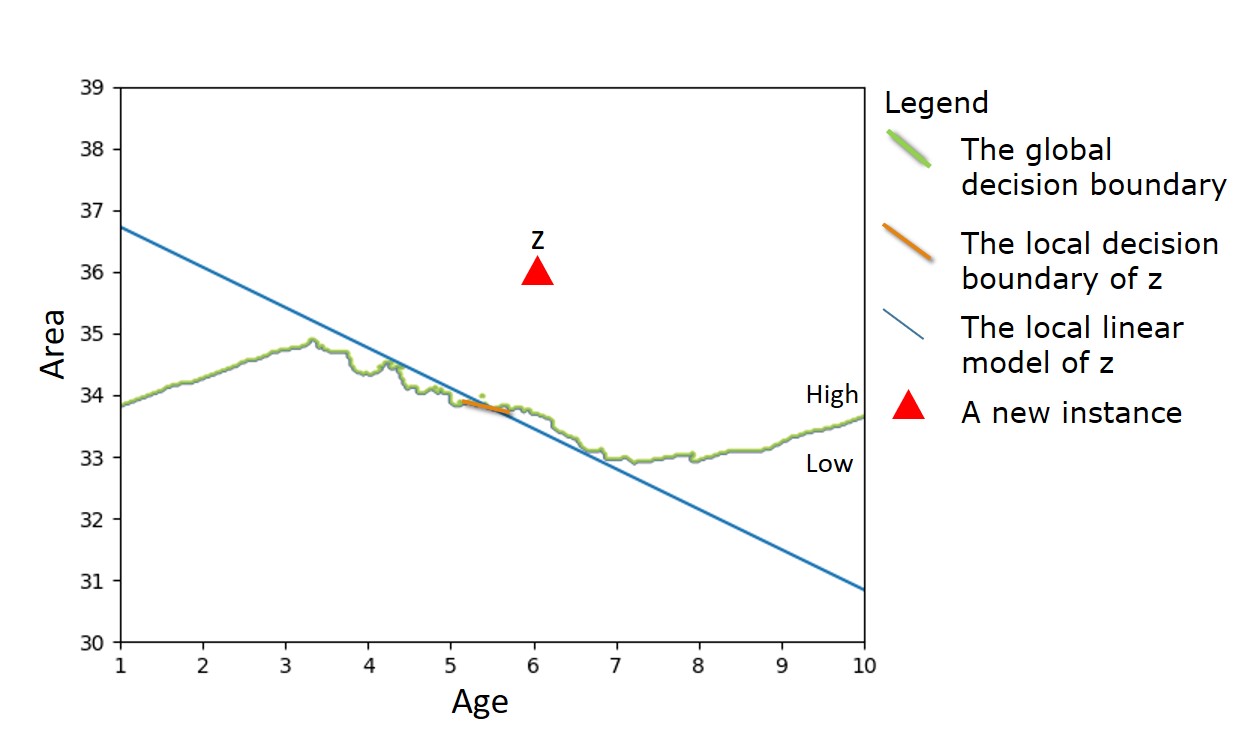}%
\caption{A complex decision boundary that cannot be accurately approximated by a linear model.}
\label{fig:complex_boundary_linear_model}
\end{center}
\end{figure}

In the simple example of Figure \ref{fig:simple_boundary_distances} the whole decision boundary can be approximated accurately by a linear model. 
ML models are usually much more complex, see e.g. Figure \ref{fig:complex_boundary_linear_model}.
However, we assume that locally the closest fragment of the global decision boundary to $\vec{z}$ is smooth enough to be linearly approximated (see the blue line in Figure \ref{fig:complex_boundary_linear_model}).



The following definitions describe the local behaviour of the ML model around $\vec{z}$.
These definitions applied to the housing example are illustrated in Figure \ref{fig:complex_boundary_linear_model}.

\theoremstyle{definition}
\begin{definition}\label{def:local_decision_boundary}
Let \emph{the local decision boundary} of $\vec{z}$ be defined as the closest fragment (according to a distance measure $D(\vec{x}_1,\vec{x}_2)$, e.g. the Euclidean distance) of the global decision boundary to $\vec{z}$.
\end{definition}

\begin{definition}\label{def:local_linear_model}
Let \emph{the local linear model} of $\vec{z}$ be the model that approximates the local decision boundary of $\vec{z}$.
\end{definition}

\begin{definition}\label{def:black_box_dissimilarity_measure}
Given the local linear model $\hat f_z(\vec{x})= \vec{w}_z \cdot \vec{x} + c$ of $\vec{z}$ let the \emph{the black-box dissimilarity measure} between $\vec{z}$ and an instance $\vec{t} \in \mathcal{X}$ be defined as the following:

\[
    b(\vec{t})= D(\vec{w}_z^T \vec{t}, \vec{w}_z^T \vec{z}) * D(\vec{t}, \vec{z}),
\]

\end{definition}

\begin{figure}[b]
\captionsetup[subfigure]{justification=centering}
\captionsetup{justification=centering}
    \centering
    \begin{subfigure}[b]{0.31\textwidth}
        \centering
        \includegraphics[width=\textwidth]{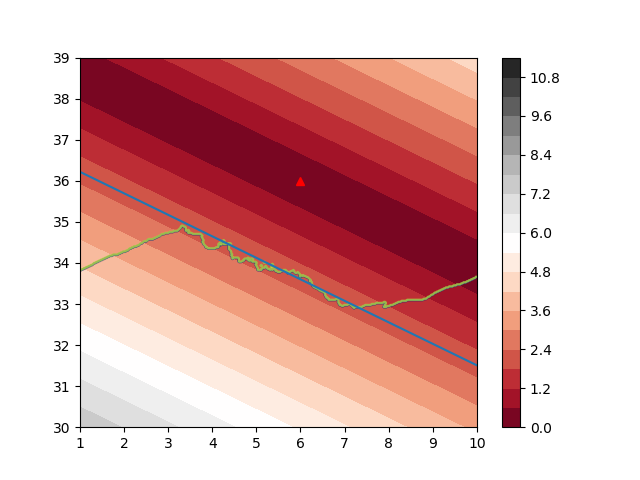}
        \caption{First factor}
        \label{fig:black_box_distance}
    \end{subfigure}
    \hfill
    \begin{subfigure}[b]{0.31\textwidth}
        \centering
        \includegraphics[width=\textwidth]{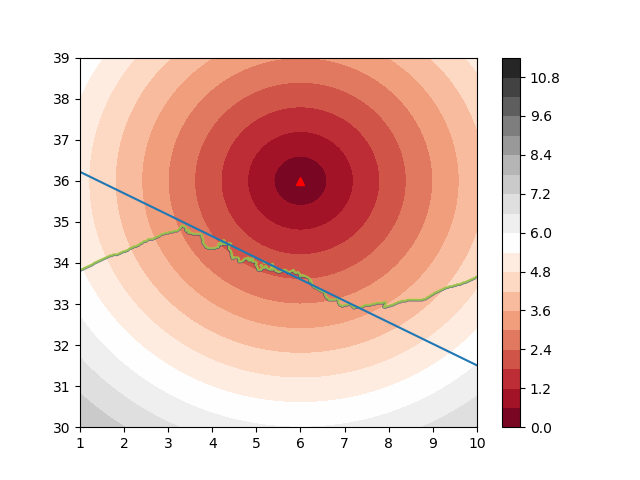}
        \caption{Second factor}
        \label{fig:euclidean_distance}
    \end{subfigure}
    \hfill
    \begin{subfigure}[b]{0.31\textwidth}
        \centering
        \includegraphics[width=\textwidth]{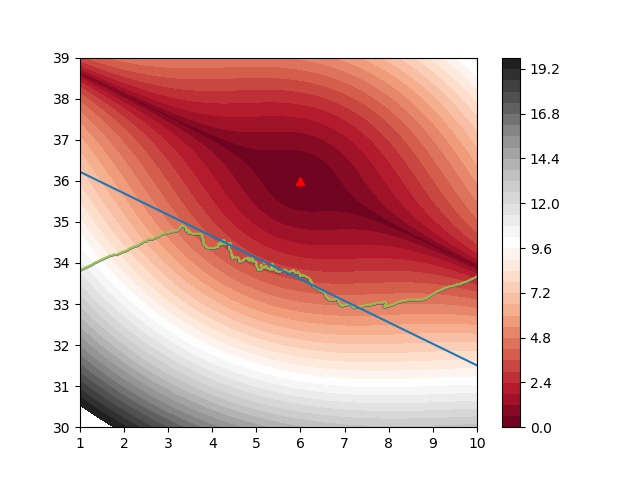}
        \caption{Whole formula}
        \label{fig:final_distance}
    \end{subfigure}
    \caption{Contour-line visualization of the black-box dissimilarity measure.}
    \label{fig:black_box_dissimilarity_measure}
\end{figure}

If $D$ is the Euclidean distance, in the 2D case the black-box dissimilarity has the form shown in Figure \ref{fig:final_distance}.
In the first factor of the Definition \ref{def:black_box_dissimilarity_measure}, $\vec{w}$ is used as weights to reflect features' importance according to $\hat f_z$ (Figure \ref{fig:black_box_distance}).
However, $\hat f_z$ is only valid in the neighbourhood $N$ of $\vec{z}$, and it is not straightforward to define $N$. 
To cope with that, we propose to leverage on the fact that closer instances to $\vec{z}$ (according to the $D$ distance measure on the input space) are more likely to be within $N$: therefore, in Definition \ref{def:black_box_dissimilarity_measure}) a second factor is added, the distance on the input space (Figure \ref{fig:euclidean_distance}).
Please note that the dissimilarity measure defined as such, does not always satisfy the condition of identity of indiscernibles $b(\vec{z'})=0 \Leftrightarrow \vec{z}=\vec{z'}$: depending on $\vec{w}_z$, the features that differ between $\vec{z}$ and $\vec{z'}$ could have no influence on $b$. Therefore this dissimilarity measure cannot be properly considered a `metric' in a mathematical sense (instead, it is a `pseudometric').

\subsection{Computation of the local linear model}\label{section:computation_local_linear_model}
The local linear model is computed from a neighborhood of $\vec{z}$ sampled from the original training set. Let us denote it as the \emph{local training set} of $\vec{z}$.

Methods to sample this local training set have been proposed in LIME \cite{ribeiro2016should} and LS \cite{laugel2018defining} (relevant details have been provided in Section \ref{section:background}). Both methods have shortcomings related to the right choice of the size of the neighbourhood from which the local training set was sampled.

Taking into account the issues of LIME and LS, we suggest two desired characteristics that a local training set of $\vec{z}$ should adhere to:

\begin{enumerate}\label{requirements}
    \item \label{itm:requirement_1} The convex hull of the local training set of $\vec{z}$ should contain the local decision boundary of $\vec{z}$.
    \item \label{itm:requirement_2} There should be enough instances to represent all classes.
\end{enumerate}

\begin{figure}
\captionsetup{justification=centering}
\begin{center}
\includegraphics[width=1\textwidth]{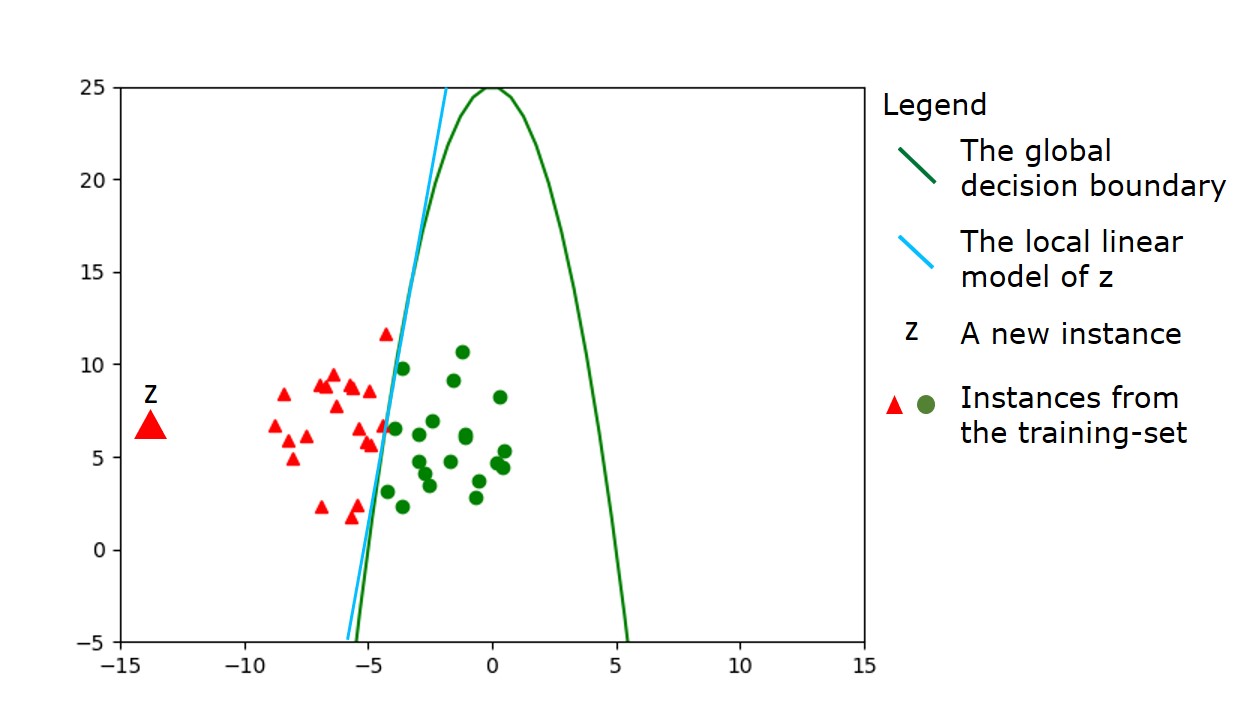}%
\caption{Sampling of the local training set of $\vec{z}$.}
\label{fig:parabola_leafage_closest_enemy}
\end{center}
\end{figure}

We propose a novel sampling strategy that covers both aspects. Its steps are:

\begin{enumerate}
    \item The local training set of $\vec{z}$ is sampled around the local decision boundary of $\vec{z}$ (similar to the idea of LS \cite{laugel2018defining}).
    This makes it possible to sample enough instances from both classes.
    We assume that the closest enemy $\vec{x}_{border}$ of $\vec{z}$ from the training set lies close to the local decision boundary of $\vec{z}$ and sample around $\vec{x}_{border}$.

    \item $i_{small} \cdot d$ samples of each class from the training set are sampled, that lie the closest to $\vec{x}_{border}$ according to the distance measure $D$.
    $d$ instances per class are the minimum amount of examples needed for a good linear approximation, assuming that these $d$ instances lie along the closest decision boundary of $\vec{z}$.
    Since these $d$ instances might not lie exactly along the decision boundary, the amount is increased with $i_{small}$ which is a small integer greater than 1.
\end{enumerate}
An example of this strategy applied on a 2D case with $i_{small}=10$ is shown on Figure \ref{fig:parabola_leafage_closest_enemy}. 
The green and red shapes are instances sampled from the training set to build the local linear model of $\vec{z}$.

Given the local training set of $\vec{z}$, a linear classification algorithm can be used to build the local linear model of $\vec{z}$.


\subsection{Explanation extraction}\label{section:explanation_extraction}
Given the local linear model $\hat f_z(\vec{x})= w_0 + w_1 x_1 + w_2 x_2 + ... + w_d x_d$ and an instance $\vec{z}=[z_1,..,z_d]$, the importance of each feature  $z_i$ can be evaluated as $abs(w_i*z_i)$, and can be provided as an explanation to the user as for which features the original model deems as relevant for its decision on $\vec{z}$. 
We refer to it as \textit{feature importance-based explanation}. 

As discussed in Section \ref{section:background}, a way to provide explanations that are closer to how humans think is to use Example-Based Reasoning, i.e. to provide examples that (according to the logic of the black-box model) are related to the test point $\vec{z}$. 
As the logic of the black-box model is locally represented by the black-box dissimilarity measure, the latter can be used to find training examples similar to $\vec{z}$ to motivate the decision.
Furthermore, one can provide both examples belonging to the predicted class $c_z$ and to the opposite class, which provides insights on the differences between classes according to the black-box model. 
We refer to them as \textit{example-based explanations}.  
Feature importance-based and example-based explanations can be also combined to provide better insights.

\begin{figure*}[b]
\captionsetup[subfigure]{justification=centering}
\captionsetup{justification=centering}
    \centering
    \begin{subfigure}[b]{1\textwidth}
        \centering
        \includegraphics[width=1\textwidth]{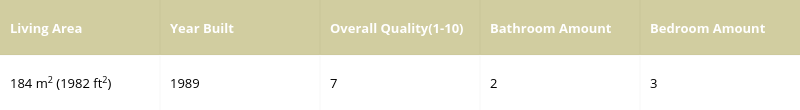}
        \caption{A house predicted as value \textit{low} by a black-box model.}
        \label{fig:101_instance_prediction_High}
    \end{subfigure}
    \hfill
    \begin{subfigure}[b]{1\textwidth}
        \centering
        \includegraphics[width=1\textwidth]{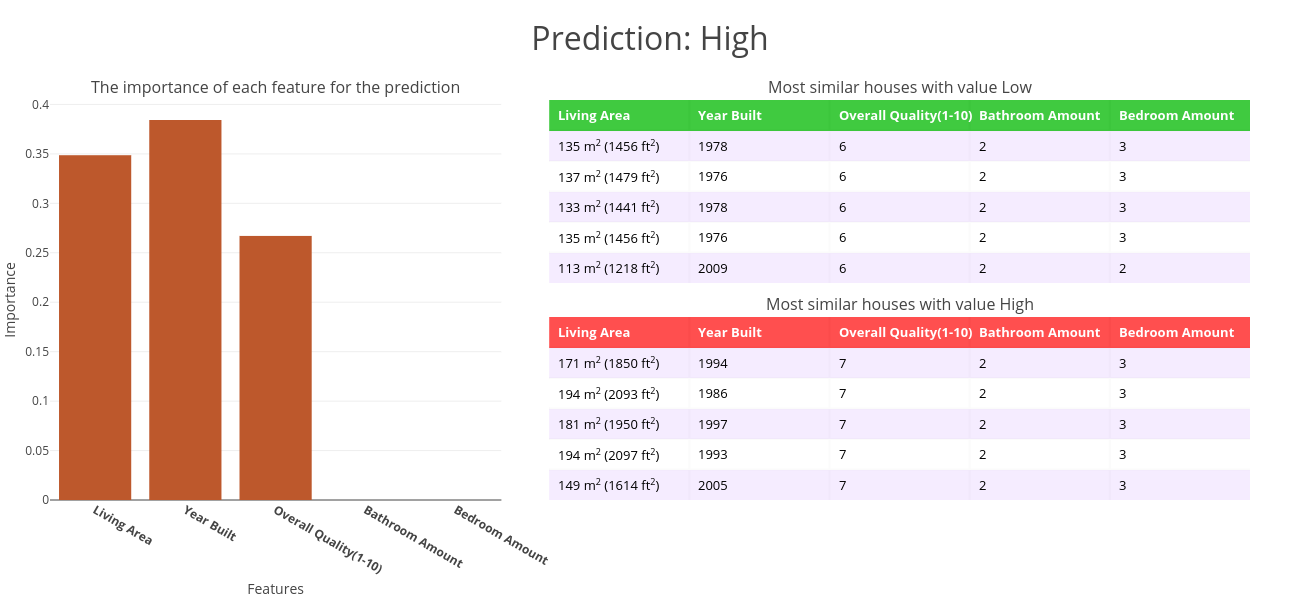}
        \caption{LEAFAGE explanation for the house above.}
        \label{fig:101_leafage}
    \end{subfigure}
    \caption{Example of a LEAFAGE explanation.}
    \label{fig:example_leafage}
\end{figure*}

An example of a test house predicted as \textit{high value} by a black-box model, and a LEAFAGE explanation for this prediction, are shown in Figures \ref{fig:101_instance_prediction_High} and \ref{fig:101_leafage}, respectively.
The left graph of Figure \ref{fig:101_leafage} shows the relative importance of each feature. 
The two tables on the right show the top 5 similar (according to the black-box dissimilarity measure) houses from the training set, belonging to the same class (\textit{high value}) and from the opposite class \textit{low value}. 
From these explanations, the user can spot insights on the classification logic, e.g., similar \textit{low value} houses have smaller \textit{living area} than similar \textit{high value} houses.

\section{Quantitative Evaluation}\label{chapter:quantitative_evaluation}
This Section evaluates the ability of LEAFAGE to reflect the true local reasoning of a black-box ML model (\textit{faithfulness} of the local approximation). 


Four different datasets with different number of features, data points and complexity are used: wine\cite{uci}, breast cancer\cite{uci}, banknote\cite{uci} and one artificial dataset.
The latter is a set of 2D data points from two highly non-separable classes. 
Instances of each class are sampled from two bi-variate normal distributions with different means ($[0, 0]$ and $[0, 1]$, respectively) and the same covariance matrix 
($\begin{bmatrix}
 2 & 0 \\
 0 & 2
\end{bmatrix}$).
The multi-class datasets are converted to binary datasets of one-vs-rest fashion.
A combination of a binary dataset and a classifier is referred to as a \emph{setting} in the following.
Each dataset is randomly split into train (70$\%$) and test set.
The train set is used to train six classifiers, namely \emph{Logistic Regression} (LR), \emph{Support Vector Machine} with linear kernel (SVM), \emph{Linear Discriminant Analysis} (LDA), \emph{Random Forest} (RF), \emph{Decision Tree} (DT) and KNN with $K=1$\footnote{\emph{scikit-learn 0.19.2} (http://scikit-learn.org/stable/) was used to build these models with their default parameters unless stated otherwise.}.
In total 36 different settings are tested.

A local linear model $\hat f_{x_i}$ is built using LEAFAGE for each instance of the test set.
Laugel. et al \cite{laugel2018defining} suggested to test the performance of $\hat f_z$ on the \textit{test} instances that fall into a hyper-sphere with a fixed radius and $\vec{z}$ as center.
Having a fixed radius has a disadvantage that the sphere may include only instances of the same class.
Therefore, we propose to use a custom radius by expanding it until the corresponding hyper-sphere includes $p$ percentage of instances that do not have the same predicted label as $c_\vec{z}$.
$p$ should be smaller than, and close to, one ($p$ = 0.95 is used in the experiments), such that the closest testing instances of the opposite class of $\vec{z}$ are included and to make the evaluation local, respectively.
The scores given by $\hat f_z$ are compared with the scores given by the black-box classifier on all the test instances that fall into this hyper-sphere, using the Area Under the ROC (AUC). 
We define the \emph{average fidelity score} as the average AUC score over the whole test set.

We then compare the average fidelity scores of LEAFAGE (with $i_{small}=10$) with LIME, in the various experimental settings as shown in Table \ref{table:fidelity_results_table_all}.

\setlength{\tabcolsep}{4pt}
\begin{table}
\tiny
\renewcommand{\arraystretch}{1.5}
\centering
\caption{Average local fidelity per setting (the standard deviation in brackets). The strategy with the highest mean along with other strategy that are statistically not significantly different are denoted in bold.}
\label{table:fidelity_results_table_all}
\begin{tabular}{cl|lll|l|l|l|}
\multicolumn{1}{l}{}                                                           & \multicolumn{1}{c|}{} & \multicolumn{3}{c|}{Wine}                                                                                                                                                            & \multicolumn{1}{l}{BreastCa.}                                  & \multicolumn{1}{l}{BankNote} & \multicolumn{1}{l}{AD}  \\
\multicolumn{1}{l}{\begin{tabular}[c]{@{}l@{}}Classifier \\Name \end{tabular}} & Strategy              & \begin{tabular}[c]{@{}l@{}}Class 0 \\vs rest \end{tabular} & \begin{tabular}[c]{@{}l@{}}Class 1 \\vs rest \end{tabular} & \begin{tabular}[c]{@{}l@{}}Class 2 \\vs rest \end{tabular} & \begin{tabular}[c]{@{}l@{}}Benign vs \\Malignant \end{tabular} & 0 vs 1                       & 0 vs 1                  \\ 
\hline
\multirow{2}{*}{LDA}                                                           & LIME                  & \textbf{100 (0.0)}                                         & \textbf{100 (0.0)}                                         & \textbf{100 (0.0)}                                         & 99.5 (1.0)                                                     & \textbf{100 (0.0)}           & \textbf{100 (0.0)}      \\
                                                                               & LEAFAGE               & 100 (0.0)                                                  & 96.0 (9.4)                                                 & 100 (0.0)                                                  & \textbf{99.9 (0.3)}                                            & 99.9 (1.7)                   & 98.6 (4.0)              \\ 
\hline
\multirow{2}{*}{LR}                                                            & LIME                  & \textbf{100 (0.0)}                                         & \textbf{100 (0.0)}                                         & \textbf{100 (0.0)}                                         & \textbf{99.9 (0.6)}                                            & \textbf{100 (0.0)}           & \textbf{100 (0.0)}      \\
                                                                               & LEAFAGE               & \textbf{100 (0.0)}                                         & \textbf{97.1 (14.2)}                                       & \textbf{100 (0.0)}                                         & 98.6 (7.8)                                                     & 99.8 (0.9)                   & 98.6 (4.0)              \\ 
\hline
\multirow{2}{*}{SVM}                                                           & LIME                  & \textbf{100 (0.0)}                                         & \textbf{100 (0.0)}                                         & \textbf{100 (0.0)}                                         & \textbf{99.9 (0.6)}                                            & \textbf{100 (0.0)}           & \textbf{100 (0.0)}      \\
                                                                               & LEAFAGE               & \textbf{100 (0.0)}                                         & \textbf{100 (0.0)}                                         & \textbf{100 (0.0)}                                         & 98.6 (7.8)                                                     & 99.8 (0.9)                   & 99.4 (1.2)              \\ 
\hline
\multirow{2}{*}{DT}                                                            & LIME                  & 91.9 (14.9)                                                & \textbf{87.9 (22.4)}                                       & 91.9 (14.7)                                                & 85.0 (16.2)                                                    & \textbf{99.0 (2.6)}          & \textbf{59.5 (32.7)}    \\
                                                                               & LEAFAGE               & \textbf{92.9 (16.0)}                                       & \textbf{85.8 (24.1)}                                       & \textbf{100 (0.0)}                                         & \textbf{86.5 (18.7)}                                           & \textbf{98.7 (4.2)}          & \textbf{65.0 (33.0)}    \\ 
\hline
\multirow{2}{*}{RF}                                                            & LIME                  & \textbf{100 (0.0)}                                         & \textbf{99.9 (0.5)}                                        & \textbf{100 (0.0)}                                         & \textbf{99.9 (0.3)}                                            & \textbf{99.1 (2.5)}          & 61.4 (36.2)             \\
                                                                               & LEAFAGE               & \textbf{100 (0.0)}                                         & \textbf{99.2 (3.7)}                                        & \textbf{100 (0.0)}                                         & \textbf{99.9 (0.8)}                                            & \textbf{98.7 (3.8)}          & \textbf{67.4 (32.9)}    \\ 
\hline
\multirow{2}{*}{KNN}                                                           & LIME                  & \textbf{98.0 (13.6)}                                       & \textbf{62.8 (37.1)}                                       & \textbf{60.5 (35.7)}                                       & 95.8 (8.2)                                                     & \textbf{100 (0.0)}           & \textbf{65.6 (34.3)}    \\
                                                                               & LEAFAGE               & 91.3 (15.9)                                                & \textbf{62.9 (36.1)}                                       & \textbf{60.3 (36.1)}                                       & \textbf{97.3 (6.0)}                                            & \textbf{99.9 (0.5)}          & \textbf{65.5 (36.8)}    \\
\hline
\end{tabular}
\end{table}

Both LIME and LEAFAGE methods perform better than a baseline model (which predicts the majority class), in all settings. 
Further, both methods work better with linear ML models (SVM, LDA, LR) as opposed as non-linear ones (DT, RF, KNN), especially with the artificial dataset. This was expected, as LEAFAGE and LIME are based on linear approximations.
On linear models, LIME scores significantly better than LEAFAGE in 11 out of 18 settings, while on non-linear models LEAFAGE performs better 5 out of 18 times.

The better performance of LIME on linear ML models could be explained by taking into account that LIME uses a high amount of samples over the whole input space to fit the local linear model.
LEAFAGE on the other hand, samples around the closest decision boundary and limits the sampling amount to a minimum (in a sense, it is more \textit{local}). This also explains the better performance of LEAFAGE over LIME on non-linear models. 

In conclusion, overall LIME performs better than LEAFAGE on linear ML models, while LEAFAGE performs better on non-linear models.

\section{Empirical Evaluation}\label{chapter:empirical_evaluation}
In order to assess the usefulness of LEAFAGE from the user perspective, we performed a user-study. 
The target group for this study was the general public. 
114 participants with a well spread demographics in term of gender, age and education (but mostly from the Americas) were recruited from Amazon Mechanical Turk, and asked to imagine they were looking for a house to buy, and that they could use an AI application to estimate the value of a house as \textit{low} or \textit{high}. 
The AI application could also provide an explanation for its estimation. 

The IOWA housing dataset \cite{de2011ames} was used, from which 5 interpretable features (i.e., features whose meaning can be directly and easily interpreted by humans) of a house were chosen, as shown in Figure \ref{fig:101_instance_prediction_High}.
We investigated 4 types of explanations for the prediction of a house, namely \textit{feature importance-based} (Figure \ref{fig:101_leafage} left), \textit{example-based} (Figure \ref{fig:101_leafage} right), a combination of \textit{example and feature importance-based} (Figure \ref{fig:101_leafage}) and\textit{ no explanation}, as a baseline.

The evaluation was split into a subjective and an objective part: \textit{perceived aid in decision-making} and \textit{objective transparency}.
In the first part, the participants were asked to rate how much they agree to the given explanation from 1 to 5 in terms of: \textit{transparency} (I understand how LEAFAGE made the prediction); \textit{information sufficiency} (the explanation provided has sufficient information to make an informed decision), \textit{competence} (the explanation corresponds to my own decision making) and \textit{confidence} (the explanation made me more confident about my decision).
Next, the objective transparency was measured by testing participants as follows: the participants were shown another house, similar to the test one; he/she had to indicate what the system would predict as the sale value of this new house.

Attention checks were implemented; Results were gathered from the 86 participants that passed the checks.

An SVM model with a RBF kernel was trained on the a training set from the IOWA dataset (70\%) to predict the binary class (\textit{low} or \textit{high} value).
All participants saw forty houses randomly chosen from the test set (30\% of the IOWA dataset), with the corresponding predicted value, and one of the four explanation types.
All participants saw the same explanations in a randomized order.
Finally, the perceived aid in decision making and objective transparency was measured.

Table \ref{table:aid_decision_making} shows the results per explanation type and dependent variable.
The median score of the explanation types differ significantly over all dependent variables according to Kruskal-Wallis H-tests \cite{kruskal1952use} with $p<0.001$ and H statistic equal to 124, 202, 55, 125 and 52 (in the left to right order of table \ref{table:aid_decision_making}, respectively).
The participants perceived getting explanation as more helpful than providing no explanation.
Dunn's post-hoc test with Bonferroni correction \cite{dunn1964multiple}  revealed that in terms of transparency, information sufficiency, competence and confidence both \textit{example-based} and \textit{combination} explanation perform significantly better than \textit{no explanation} and \textit{feature importance-based} explanation, while no significant differences were found between \textit{example-based} and \textit{combination} explanations.
Moreover, \textit{feature importance-based} explanation performed significantly better than \textit{no explanation} regarding transparency, information sufficiency and confidence but not in terms of competence.
However, in terms of objective transparency, \textit{feature importance-based} explanation performed significantly worse than the rest of the explanation types including \textit{no explanation}.
\textit{Example-based} explanation has the highest average objective transparency score, however no statistically significant difference was measured between pairs of \textit{no explanation}, \textit{example-based} and \textit{combination} explanation.

Finally, the participants provided general remarks for each explanation type.
When no explanation was provided, they indicated that they could still understand the prediction, but that they needed ``complete trust in the system to find it helpful".
The participants liked the simplicity and visual aspect of the \textit{feature importance-based} explanation, but they did not find it detailed enough to perform well on the objective transparency part.
Moreover, they found it hard to estimate what value of a certain feature changes the prediction, and how the importance really relates to the prediction of a house.
Regarding \textit{example-based} explanation, participants appreciated that they could compare similar houses with different sale values.
However, some participants disliked this explanation type because of the amount of information present in the tables.
Finally, the combination of example-based and feature importance-based explanation received a mixed reaction: some participants liked to get a detailed explanation while others were overwhelmed and focused on one chart and ignored the other.

\setlength{\tabcolsep}{6pt}
\begin{table}[]
\tiny
\begin{tabular}{@{}lllllll@{}}
\toprule
                         & Transparency & Info. Suff. & Competence  & Confidence  & Objec. Trans.\\ \midrule
No Explanation           & 3.66 (1.03)  & 3.43 (1.17)       & 3.70 (0.96) & 3.52 (1.1)  & 8.40 (1.48)\\
Feature importance       & 3.92 (0.85)  & 3.76 (0.97)       & 3.78 (0.91) & 3.68 (1.04) & 7.20 (1.66)\\
Example-based            & 4.07 (0.76)  & 4.02 (0.84)       & \textbf{3.96} (0.86) & \textbf{3.98} (0.86) & \textbf{8.83} (1.40)\\
Ex. and Feat.            & \textbf{4.13} (0.8)   & \textbf{4.10} (0.83) & 3.93 (0.93) & \textbf{3.98} (0.9)  & 8.56 (1.68)\\ \bottomrule
\end{tabular}
\caption{Results of perceived aid in decision making and objective transparency per explanation type.}
\label{table:aid_decision_making}
\end{table}

\section{Conclusion}\label{chapter:conclusion}
In this paper, we presented LEAFAGE, a novel method to provide local explanations for the predictions of a black-box ML model.
LEAFAGE explanations performed overall better than the state of the art on non-linear models in terms of local fidelity. 
We also evaluated LEAFAGE empirically, by engaging people in a user study. This aspect has been largely overlooked by the scientific literature on XAI.

The empirical evaluation showed that overall the participants perceived having explanations behind a prediction as more helpful than having no explanation for the goal of decision making.
Interestingly, when participants were tested about their gained knowledge after seeing an explanation, no significant advantage was found compared to providing no explanation.
We suspect that this is due to the simplicity of the test, in future work a more comprehensive test could be used to measure the actual transparency.
The user study also showed that, with regards to objective transparency, feature importance-based explanations are less effective than providing no explanation at all.
This is an important result, which suggests that feature importance-based explanation confuses users more about the prediction than providing no explanation.

Further, example-based explanations performed significantly better than feature-importance based explanation in terms of perceived aid in decision making.
Showing both example-based and feature-importance-based explanation did not increase the perceived aid in decision making significantly.
This could be due to the overload of information as the participants described.
Interestingly, some participants indicated that a tabular view of the example-based explanation was hard to read. In our future work, we will focus on designing them in a more readable and intuitive manner.

\section*{Acknowledgement}
This research is supported by the Hybrid AI Explainability and VP AI \& Robotics programs at the Netherlands Organisation for Applied Scientific Research (TNO).

\bibliographystyle{unsrt}
\bibliography{main}

\end{document}